# Aspect-based Opinion Summarization with Convolutional Neural Networks

**Haibing Wu, Yiwei Gu, Shangdi Sun** and **Xiaodong Gu**
Department of Electronic Engineering
Fudan University
Shanghai 200433, China
`{haibingwu13, yiweigu11, shangdisun14, xdgu}@fudan.edu.cn`

## Abstract

This paper considers Aspect-based Opinion Summarization (AOS) of reviews on particular products. To enable real applications, an AOS system needs to address two core subtasks, aspect extraction and sentiment classification. Most existing approaches to aspect extraction, which use linguistic analysis or topic modelling, are general across different products but not precise enough or suitable for particular products. Instead we take a less general but more precise scheme, directly mapping each review sentence into pre-defined aspects. To tackle aspect mapping and sentiment classification, we propose two Convolutional Neural Network (CNN) based methods, cascaded CNN and multitask CNN. Cascaded CNN contains two levels of convolutional networks. Multiple CNNs at level 1 deal with aspect mapping task, and a single CNN at level 2 deals with sentiment classification. Multitask CNN also contains multiple aspect CNNs and a sentiment CNN, but different networks share the same word embeddings. Experimental results indicate that both cascaded and multitask CNNs outperform SVM-based methods by large margins. Multitask CNN generally performs better than cascaded CNN.

## 1 Introduction

User generated reviews on products are expanding rapidly with the emergence and advancement of e-commerce. These reviews are valuable to business organizations for improving their products and to individual consumers for making informed decisions. Unfortunately, reading though all the product reviews is hard, especially for popular products with large volume of review texts. It is t-herefore essential to provide coherent and concise summaries of user generated reviews. This has bred a new line of research on Aspect-based Opinion Summarization (AOS) (Hu and Liu, 2004). Given a set of product reviews, an AOS system extracts aspects discussed in the reviews and predicts reviewers' sentiments toward these aspects. Figure 1 presents an example summary of smartphone reviews. The smartphone aspects, such as battery life and screen, with the hyperlinks and numbers of positive and negative opinions, are illustrated in a structured way.

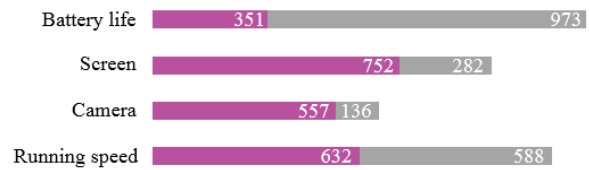

Figure 1. An example aspect-based summary of smartphone reviews.

Standard AOS typically involves two component subtasks, aspect extraction and sentiment classification. Aspect extraction finds related aspects and extracts all textual mentions associated with each aspect. Sentiment classification task classifies sentiment over each aspect using the associated textual mentions.

Existing researches on aspect extraction move along two quite different lines. The first extracts aspect expressions using linguistic patterns or supervised sequence labeling (see Section 2). This scheme is very limited for only identifying explicit aspects and failing to handle implicit aspects. Besides, it needs additional efforts to group synonymous aspect expressions into the same category. The second is based on topic modeling (see Section 2). Topic modeling is fully unsupervised, saving the labeling of training data. It handles implicit aspects well, and simultaneously extracts and groups aspects. It is, however, not suitable for summarizing reviews on particular products in many respects. The unsupervised nature makes it more general across different products, but less precise for particular products compared to supervised learning methods. The learned topics of topic modeling are implicit and often do not correlate well with human judgments, making it not applicable if users care about some particular product aspects. Topic modeling categorizes aspects, but its unsupervised nature makes the grouping not controllable or adaptable. Categorizing aspects is subjective because for different applications the user may need different categorizations. For example, in smartphone reviews, front camera and back camera can be regarded as two separate aspects, but can also be one general aspect, camera.

For some vertical e-commerce websites that focus on particular products, users already know what aspects a product has. Ontologies negates the need for identifying aspects automatically. Herein the most pressing challenge is to extract all relevant text mentions for each aspect. Therefore, this paper takes a line different from prior work on aspect extraction: directly mapping each review sentence into pre-defined aspect categories. That is, we formulate aspect extraction as sentence-level aspect mapping (or classification) problem. This scheme extracts relevant text mentions for pre-defined aspects and enjoys a lot of advantages. It handles both explicit and implicit aspects, and simultaneously extracts and categorizes different aspect expressions into the same aspect category. It also enables users to design different aspect categories for different application purposes.

Besides aspect extraction, sentiment classification is also necessary to enable real applications. This paper presents an aspect-based summary system which addresses both tasks. Most previous work on AOS deals with a single task, either aspect extraction or sentiment classification, using traditional machine learning. Motivated by the recent success of deep Convolutional Neural Network (CNN) and multi-task representation learning, we propose two CNN-based approaches to jointly tackle aspect mapping and sentiment classification problems. The first one is a two-level Cascaded CNN (C-CNN). At level 1, multiple convolutional networks map the input sentences into pre-defined aspects. At level 2, a single convolutional network predicts the sentiment polarities of the input sentences. The second is a Multi-task CNN (M-CNN). Different from C-CNN, aspect mapping and sentiment classification tasks share the word embedding representation in MT-CNN, making the learned word embedding universal across tasks. This reduces over-fitting to a specific task and thus profits generalization to held-out test data. Empirical results show that both C-CNN and M-CNN with pre-trained word embedding representation outperform linear classifiers with bag-of-words representation by large margins. M-CNN performs better than C-CNN, despite not showing significant superiority.

## 2 Related Work

AOS has attracted a lot of attentions with the advent of online user generated reviews. Deep learning and representation learning, initially enjoying great success in computer vision, have also achieved some success in Natural Language Processing (NLP).

An AOS system needs to address two core tasks, aspect extraction or sentiment classification. One line of work on aspect extraction detects aspect expressions using linguistic patterns (e.g. part-of-speech and dependency relations) (Hu and Liu, 2004; Joshi and Rose, 2009; Zhuang et al., 2006; Wu et al., 2009; Qiu et al., 2011) or supervised sequence labeling such as CRFs (Jin and Hung, 2009; Jakob and Gurevych, 2010; Irsoy and Cardie, 2014; Breck et al., 2007; Yang and Cardie, 2012). This scheme is very limited in many respects. It only extracts explicit aspect expressions, and cannot deal with implicit aspects well. For example, in the sentence "this phone runs smoothly and fast, but its battery life is very poor", *battery life* is explicitly mentioned, while *running speed* is implicitly mentioned and thus cannot directly discovered using linguistic patterns or sequence labeling. This scheme is also limited for not grouping aspect expressions into aspect categories. For example, *screen*, *display* and *retina* refer to the same aspect for iPhone. After extracting all aspect expressions, additional efforts are required to categorize domain synonyms into the same aspect.

Another line of related work applies variants of standard topic modeling such as LDA (Titov and McDonald, 2008; Christina et al., 2011; Brody and Elhadad, 2010; Zhao et al., 2010; Jo and Oh, 2011; Moghaddam and Ester, 2012; Chen et al., 2014; Mukherjee and Liu, 2012; Kim et al., 2013; Sauper and Barzilay, 2013). Topic modeling deals with implicit aspects to some degree, and simulta-neously extracts and groups aspects. However, it often learns incoherent topics since its objective functions do not always correlate well with human judgments. Compared with supervised methods, unsupervised topic modeling is more general across different products, but less precise for particular products. In addition, mapping from topics to aspects is not explicit, making it not a good choice if users care about opinions on some particular aspects. Topic modeling categories aspects based on co-occurrence counts. However, categorizing aspects is subjective because for different applications the user may need different categorizations. For example, in smartphone reviews, front camera and back camera can be treated as two different aspects, but can also be only one aspect. The unsupervised nature of topic modeling makes the grouping not controllable or adaptable.

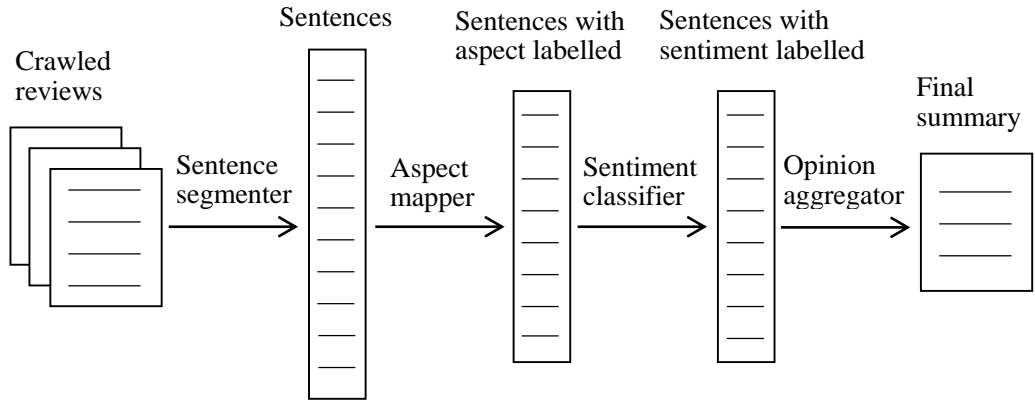

Figure 2. An overview of our aspect-based opinion summarization system.

An AOS system also involves sentiment classification. This task aims to classify an opinionated review as expressing positive or negative sentiment over an aspect. Compared to aspect extraction, sentiment classification was studied earlier and more extensively. Most prior work used traditional machine learning with complicated feature engineering (Pang et al., 2002; Ng et al., 2006; Riloff et al., 2006; Davidov et al., 2010; Paltoglou and Thelwall, 2010; Nakagawa et al., 2010; Bespalov et al., 2011; Wu and Gu, 2014). Very recently, some researchers applied deep convolutional neural networks to sentence sentiment classification and reported considerably better results than traditional approaches (Kalchbrenner et al., 2014; Santos and Gatti, 2014; Kim, 2014).

Convolutional neural network (CNN) is currently underpinning the cutting edge in computer vision (Krizhevsky et al., 2012; Szegedy et al., 2014). It has also achieved state-of-the-art results in many traditional NLP tasks (Collobert et al., 2011) and other NLP areas such as information retrieval (Shen et al., 2014) and relation classification (Zeng et al., 2014; Santos et al., 2015). Words are encoded as low-dimensional word vectors in CNN, instead of high dimensional one-hot representations. Word vector representations capture semantic information, so semantically close words are likewise close in low dimensional vector space. CNN models for specific NLP tasks often use unsupervised pre-trained word vectors (Mikolov et al., 2013) as initialization, which are then improved by optimizing supervised objectives.

Multi-task learning learns shared representation for multiple tasks. In NLP, the marriage between multi-task learning and neural networks is quite natural as different NLP tasks could share word embeddings. For example, (Collobert et al., 2011) tackled part-of-speech tagging, chunking, and named entity recognition tasks using a multi-task sequence labeller. (Liu et al., 2015) trained a multi-task deep neural network for query classification and web search ranking.

## 3 Methodology

An architectural overview of our aspect-based summarization system is given in figure 2. The input to the system is a set of crawled reviews for a particular product. The sentence segmenter divides review texts into a set of sentences. The aspect mapper maps these sentences into pre-defined aspect categories. In this step only sentences belonging to the pre-defined aspects are extracted and retained. The sentiment classifier then predicts the polarity of each of these extracted sentences as positive or negative. After labelling each sentence with aspect and sentiment, the final opinion aggregator counts the number of positive and negative opinionated sentences corresponding to each aspect, and gives the hyperlinks to these sentences.

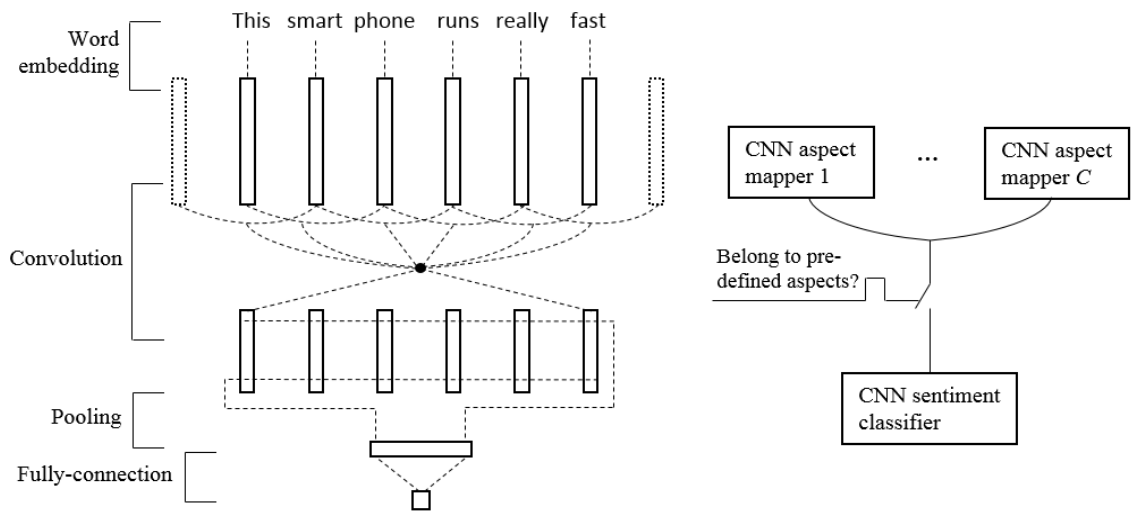

Figure 3. Our C-CNN for aspect mapping and sentiment classification.

The sentence segmenter in our system is an off-the-shelf segmentation tool, NLTK PunktSentenceTokenizer.[1] The aspect mapper and sentiment classifier we use is a C-CNN or M-CNN. For each input sentence, the network first maps it into corresponding aspects. If the sentence belongs to one or more pre-defined aspect categories, the network then predicts its sentiment polarity.

### 3.1 Cascaded CNN

The architecture of our C-CNN is shown is figure 3. The network contains $C$ CNN aspect mappers and a CNN sentiment classifier. Aspect-mapping CNN and sentiment-classification CNN are organized in a cascaded way. Each mapper determines whether the input sentence belongs to its corresponding aspect. If that is the case, the sentiment classifier predicts sentiment polarity as positive or negative.

Before diving into details about CNN layers, we address two considerations about the cascaded network. (1) The network only contains one sentiment classifier. One may think it is problematic as a single sentence (e.g. "This phone runs fast, but does not loses its charge too quickly!") could contain different aspects, and sentiments towards these aspects could be opposite. We do not train a separate sentiment classifier for each aspect category since in practice only a few sentences imply opposite sentiments for different aspects. (2) The sentiment classifier only deals with sentences belonging to at least one pre-defined aspects as practical applications only care the sentiments of aspect related sentences. In addition, sentences not belonging to any pre-defined aspect could be objective. It is not suitable to classifying the sentiments of objective sentences as positive or negative.

Each CNN contains a word embedding layer, a convolutional and pooling layer, and a fully-connected layer.

**Word embedding.** This layer encodes each word in the input sentence as a word vector. Let $l \in R$ be the sentence length, $|D| \in R$ be the vocabulary size and $W^{(1)} \in R^{k \times |D|}$ be the embedding matrix of $k$-dimensional word vectors. The $i$-th word in a sentence is transformed into a $k$-dimensional vector $w_i$ by matrix-vector product:

$$w_i = W^{(1)} x_i. \qquad (1)$$

Here $x_i$ is the one-hot representation for the $i$-th word.

**Convolution.** After encoding the input sentence with word vectors, the convolution operations are applied on top of these vectors to produce new features. A convolution operation involves a filter $u \in R^{hk}$ applied to a window of $h = 2r+1$ words. For example, a feature $f_i$ is produced from a window of words $w_{i-r:i+r}$ by

$$f_i = g(w_{i-r:i+r} \cdot u). \qquad (2)$$

Here $g$ denotes a non-linear activation function. This filter is applied to every possible windows of the input sentence to generate a feature map[2]

---

[1] http://www.nltk.org/api/nltk.tokenize.html

[2] Each sentence is augmented with $r$ "padding" words respectively at the beginning and the end, so each word in a sentence corresponds to a convolution window.

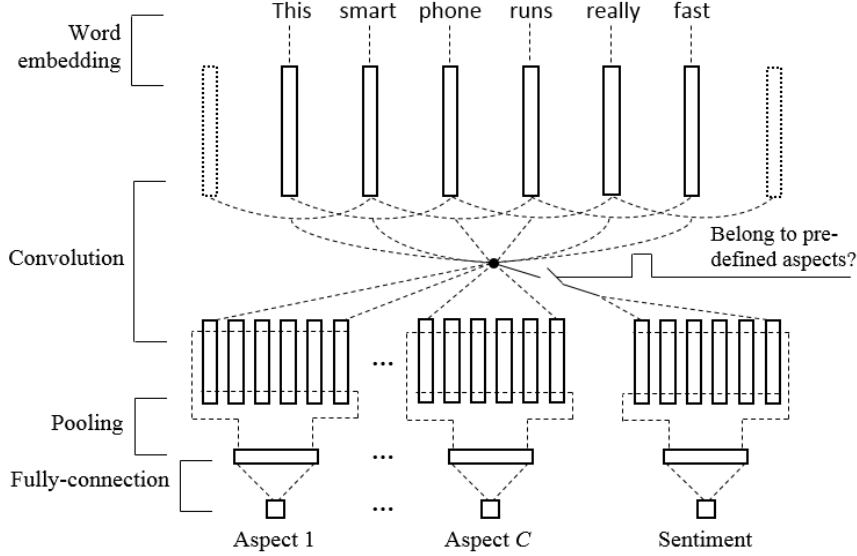

Figure 4. Our M-CNN architecture for aspect mapping and sentiment classification.

$$f = [f_1, f_2, ..., f_l]. \quad (3)$$

The above describes the process that one feature map is extracted from one filter. The network uses $m_i$ ($i = 1, 2, ..., C$) filters to generate $m_i$ feature maps for the $i$-th aspect mapper and $m_{C+1}$ filters for the sentiment classifier. The filter weights for $i$-th aspect mapper are stored in a $hk \times m_i$-dimensional matrix $W_i^{(2)} \in R^{hk \times m_i}$. For sentiment classifier, $W_{C+1}^{(2)} \in R^{hk \times m_2}$.

**Pooling.** This layer applies max-over-time pooling (Collobert et al. 2011) to each of the feature maps produced by convolutional layers:

$$\hat{f} = \max(f_1, f_2, ..., f_l). \quad (4)$$

Max-over-time pooling takes the maximum element in each feature map and naturally deals with variable sentence lengths. It produces a fixed-sized feature vector $v_i \in R^{m_i}$ for the $i$-th task.

**Fully-connection.** The fixed-sized feature vectors produced by pooling layers are fed into fully-connected layers. Concretely, $v_i$ is passed to a binary logistic regression classifier.

$$a_1 = 1 / (1 + e^{-W_i^{(3)} v_i}), i = 1, 2, ..., C+1. \quad (5)$$

Here $W_i^{(3)} \in R^{n \times m_i}$ is the weight matrix for $i$-th task, and $a_i$ is the aspect output vector. For aspect mapper, $a_i$ ($i = 1, 2, ..., C$) is the probability of the input sentence belonging to the $i$-th aspect category; for sentiment classifier $a_i$ ($i = C+1$) is the positive-sentiment probability.

### 3.2 Multitask CNN

The architecture of our designed M-CNN is shown is figure 4. M-CNN also contains $C$ aspect mappers and a sentiment classifier. But different to C-CNN, aspect mappers and sentiment classifier share word embedding layer in M-CNN. So the word embedding parameter $W^{(1)}$ is shared across different tasks, whereas other parameters, i.e. $W_i^{(2)}$ and $W_i^{(3)}$ ($i = 1, 2, ..., C + 1$), are task specific.

Conventional multitask learning optimizes model parameters $\theta = (W^{(1)}, W^{(2)}, W^{(3)})$ by minimizing the loss functions of all tasks. This gives slightly worse results in our experiments. Instead we sequentially set task $i$ as the main task, and set the other tasks as auxiliary tasks. The aim is to optimize the main task, with the assistance of auxiliary tasks. To this end, we formulate the optimization objective as

$$J(\theta) = \ell_i + \sum_{j \neq i} \lambda_j \ell_j, \quad (6)$$

where $\ell_i$ is loss function of the main task, and $\ell_j$ is the loss function of each auxiliary task. The parameter $\lambda_j$ ($0 < \lambda_j < 1$) denotes the importance of $j$-th task. Logistic loss is used as loss functions for both aspect mapping and sentiment classification tasks.

In order to optimize $J(\theta)$ we use mini-batch stochastic gradient descent as shown in Algorithm 1. The algorithm sequentially set task $i$ as the main task, set and other tasks as auxiliary tasks. It then updates $W^{(1)}$, $W_i^{(2)}$ and $W_i^{(3)}$ for the main task and all auxiliary tasks using gradient descent. After training parameters for $T$ epochs, the learned model is test on the held-out test sentences for task $i$.

Note that if a task is sentiment classification, gradients are only computed over sentences belo-ging to at least one aspect category because sentences not belonging to any pre-defined aspect are filtered out.

**Algorithm 1:** Training and testing our M-CNN
**for** $i = 1$ to $C + 1$
1. set task $i$ as the main task, and set other tasks as auxiliary tasks
2. repeat step 3-10 for $T$ epochs
3. permute training sentences randomly and partition them into mini-batches
4. **for** each mini-batch
5.   compute gradients of $J(\theta)$ w.r.t. $W^{(1)}$, $W_i^{(2)}$ and $W_i^{(3)}$ for the main task
6.   update $W^{(1)}$, $W_i^{(2)}$ and $W_i^{(3)}$ using gradient descent
7.   compute gradients of $J(\theta)$ w.r.t. $W^{(1)}$, $W_j^{(2)}$ and $W_j^{(3)}$ ($j \neq i$) for all auxiliary tasks
9.   update $W^{(1)}$, $W_j^{(2)}$ and $W_j^{(3)}$ ($j \neq i$) using gradient descent
10. **end**
11. test the learned model on the test sentences for task $i$
**end**

| Aspects | #Sentences |
|---|---|
| battery | 352 |
| run speed | 370 |
| speaker | 158 |
| screen | 434 |
| camera | 344 |
| others | 11,042 |
| all | 12,700 |

Table 1. The number of sentences belonging to each aspect category.

## 4 Experiments

**Dataset.** To train our C-CNN and M-CNN, we need a collection of sentences labeled with aspects and sentiments. As there is no such benchmark corpus, we create an Amazon Smartphone Review (ASR) dataset and will make it publicly available for research purpose. ASR contains 300,000 smartphone reviews crawled from amazon.com. We manually annotate 12,700 sentences of 1679 reviews with respect to five pre-defined aspects, {battery, screen, camera, speaker, running speed}. Sentences belonging to at least one aspect are also labeled as expressing positive or negative sentiment. The number of sentences belonging to each aspect is shown in table 1.

**Evaluation metrics.** We use precision, recall and F1-score for performance evaluation of aspect mapping, and classification accuracy for sentiment classification. All comparisons are done using 5-fold cross validation. That is, the overall results are averaged over five folds.

**Baselines.** The baselines exploit Support Vector Machines (SVMs) as classifiers. Specially, we adopt the L2-regularized L2-loss linear SVM and the implementation software is scikit-learn.[3] Multiple SVMs are cascaded in the way like C-CNN. One-hot representation of each word (or term) is employed as feature for training SVMs. The weight of each word in the one-hot representation is simply assigned term presence (tp), i.e. 1 for presence and 0 for absence. The most commonly used weighting scheme, term frequency (tf), is not used as it produce very close results to tp. The reason may be that in our experiments most words in a sentence only occur one time, so weights assigned by tp and tf are almost the same with each other.

**Network settings.** We use rectified linear units (Nair and Hinton, 2010) as activation functions for convolutional layer, and sigmoid function for output layer. Network models are trained using stochastic mini-batch gradient descent with batch size of 1000, momentum of 0.9, learning rate of 0.5. The weights in all layers are initialized from a zero-mean Gaussian distribution with 0.1 as standard deviation and the constant 0 as the neuron biases. For M-CNN, the parameter $\lambda_j$ is manually chosen according to the performance on development set. The settings of hyper-parameters on the architectures of C-CNN and M-CNN are shown in table 2.

**word2vec (w2v).** Besides random initialization, we also pre-train word embeddings using word2vec tool,[4] which implements continuous bag-of-words and skip-gram architectures for learning word vector representations (Mikolov et al., 2013). We train skip-gram model with context window size of 9 on corpus of December 2013 English Wikipedia.

---
[3] scikit-learn.org/
[4] code.google.com/p/word2vec/

| Hyper-parameter | Description | Aspect mapping | Sentiment classification |
|---|---|---|---|
| $k$ | Word embedding dimension | 30 | 30 |
| $h$ | Convolution window size | 3 | 3 |
| $m_i$ | Number of filters | 300 | 100 |

Table 2: Network hyper parameters.

| | battery | | | run speed | | | speaker | | | screen | | | camera | | |
|---|---|---|---|---|---|---|---|---|---|---|---|---|---|---|---|
| Methods | P | R | F | P | R | F | P | R | F | P | R | F | P | R | F |
| SVM | 72.4 | 70.9 | 71.7 | 72.8 | 64.0 | 68.1 | 81.6 | 64.5 | 72.1 | 74.9 | 71.5 | 73.2 | 80.9 | 78.0 | 79.4 |
| C-CNN | 76.1 | 72.8 | 74.4 | 77.1 | 63.5 | 69.6 | 86.4 | 64.4 | 73.8 | 78.9 | 72.5 | 75.6 | 81.9 | 77.9 | 79.9 |
| M-CNN | 77.2 | 72.6 | 74.8 | 77.9 | 64.2 | 70.4 | 86.2 | 65.3 | 74.3 | 79.1 | 72.0 | 75.3 | 81.7 | 78.5 | 80.1 |
| C-CNN + w2v | 77.8 | 74.4 | 76.1 | 78.1 | 64.7 | 70.8 | **87.1** | 65.2 | 74.6 | **79.7** | 73.1 | **76.3** | **82.9** | 79.1 | 80.9 |
| M-CNN + w2v | **78.2** | **74.7** | **76.4** | **78.9** | **65.6** | **71.6** | 86.8 | **66.4** | **75.2** | 79.3 | **73.2** | 76.1 | 82.8 | **79.7** | **81.2** |

Table 3. Precision, recall and F1-score of different methods for aspect mapping.

### 4.1 Results for Aspect Mapping

Table 3 presents the results of our CNN-based methods against SVM-based methods for aspect mapping. Clearly, our C-CNN and M-CNN with randomly initialized word embeddings perform better than SVM with tp for all five aspects. For example, comparing C-CNN to SVM, the increases of F1-score are respectively 2.7% (71.7% vs. 74.4%), 1.5% (68.1% vs. 69.6%), 1.7% (72.1% vs. 73.8%), 2.4 % (73.2% 75.6%) and 0.5% (79.4% vs. 79.9%). In terms of precision, CNN methods also beat SVMs with large margins, while their recall performances are close.

M-CNN generally outperforms C-CNN, but does not show significant superiority. For the aspects battery, running speed, speaker and camera, M-CNN produces higher F1-score than C-CNN, but for screen M-CNN slightly underperforms C-CNN (75.3% vs. 75.6%). Overall, the performance between M-CNN and C-CNN is close to each other, with the former performing slightly better than the latter.

Pre-training word embeddings using word2vec provides significant gains of precision, recall and F1-score for C-CNN and M-CNN on all aspect mapping tasks. The improvement of F1-score ranges from +0.7% (75.6% vs. 76.3%) to +1.7% (74.4% vs. 76.1%). We also observe the evidence of the benefits of multitask learning after using word2vec as word embedding initializer. In terms of F1-score, M-CNN+w2v outperforms C-CNN+w2v on 4 of 5 aspects.

### 4.2 Results for Sentiment Classification

Table 4 presents the classification accuracy of our CNN-based methods against SVM-based methods for sentiment classification task. Different methods' performances for this task are generally in accordance to their performances for aspect mapping tasks. SVM with tp produces an accuracy of 80.3%. Again, both C-CNN and M-CNN outperform SVM+tp by large margins. The accuracy differences are respectively +2.1 (80.3 % vs. 82.4%) and 2.6% (80.3% vs. 82.9%). C-CNN+w2v and M-CNN+w2v achieve accuracy of 83.5% (vs. 82.4% by C-CNN) and 84.1% (vs. 82.9% by M-CNN) respectively, implying the power of unsupervised pre-training of word embeddings. M-CNN+w2v achieves the highest accuracy among all evaluated methods.

| Methods | Accuracy |
|---|---|
| SVM | 80.3 |
| C-CNN | 82.4 |
| M-CNN | 82.9 |
| C-CNN + w2v | 83.5 |
| M-CNN + w2v | **84.1** |

Table 4. Classification accuracy of different methods for sentiment classification.

## 5 Conclusions

In this paper we have presented an aspect-based opinion summarization system for particular products. Our system directly maps each review sentence into pre-defined aspects. This is particularly suitable for some vertical e-commerce websites that only sell particular products, or if users only care about opinions on particular product aspects. To attack aspect mapping and sentiment classification tasks, we have proposed two convolutional network based approaches, C-CNN and M-CNN. Both C-CNN and M-CNN contains multiple aspect mappers and a single sentiment classifier. The difference is that in M-CNN, word embeddings are shared across different tasks. Empirical results imply the superiority of our CNN-based methods over SVM-based methods; M-CNN tends to perform better than C-CNN, despite not showing significant superiority.

## Acknowledgements

This work was supported in part by National Natural Science Foundation of China under grant 61371148 and Shanghai National Natural Science Foundation under grant 12ZR1402500.